\documentclass{ecai} 

\usepackage{latexsym}
\usepackage{amssymb}
\usepackage{amsmath}
\usepackage{amsthm}
\usepackage{svg}
\usepackage{booktabs}
\usepackage{enumitem}
\usepackage{graphicx}
\usepackage{color}
\usepackage{xcolor}
\usepackage{multirow}
\usepackage{booktabs}
\usepackage[group-separator={,}]{siunitx}
\usepackage{soul,color}

\usepackage{hyperref}
\hypersetup{
  colorlinks = true, 
  linkcolor = black, 
  anchorcolor = black, 
  citecolor = black, 
  filecolor = black, 
  menucolor = black, 
  urlcolor = black, 
}

\newcommand{\BibTeX}{B\kern-.05em{\sc i\kern-.025em b}\kern-.08em\TeX}

\newcommand{\alvaro}[1]{{\color[HTML]{000000}{#1}}}

\newcommand{\javi}[1]{{\color[HTML]{000000}{#1}}}

\DeclareGraphicsExtensions{.pdf,.png,.jpg,.svg}

\newcommand{\DIFdel}[1]{{\color{green}\sout{}}} 

\usepackage[normalem]{ulem} 

\usepackage{orcidlink} 

\begin{document}


\begin{frontmatter}


\paperid{123} 


\title{Uncertainty Quantification for Transformer Models for Dark-Pattern Detection}


 \author[A]{\fnms{Javier}~\snm{Muñoz}\orcidlink{0000-0001-5068-3303}\thanks{Corresponding Author. Email: jmunoz@funditec.es.}}
 \author[A]{\fnms{Álvaro}~\snm{Huertas-García}\orcidlink{0000-0003-2165-0144}}
 \author[A]{\fnms{Carlos}~\snm{Martí-González}\orcidlink{0009-0003-1387-1630}} 
 \author[A]{\fnms{Enrique}~\snm{De Miguel Ambite}}

 \address[A]{Advantx Technological Foundation (Funditec), Madrid, Spain}






\begin{abstract}
The opaque nature of transformer-based models, particularly in applications susceptible to unethical practices such as dark-patterns in user interfaces, requires models that integrate uncertainty quantification to enhance trust in predictions. This study focuses on dark-pattern detection, deceptive design choices that manipulate user decisions, undermining autonomy and consent. We propose a differential fine-tuning approach implemented at the final classification head via uncertainty quantification with transformer-based pre-trained models.
Employing a dense neural network (DNN) head architecture as a baseline, we examine two methods capable of quantifying uncertainty: Spectral-normalized Neural Gaussian Processes (SNGPs) and Bayesian Neural Networks (BNNs). These methods are evaluated on a set of open-source foundational models across multiple dimensions: model performance, variance in certainty of predictions and environmental impact during training and inference phases. Results demonstrate that integrating uncertainty quantification maintains performance while providing insights into challenging instances within the models. Moreover, the study reveals that the environmental impact does not uniformly increase with the incorporation of uncertainty quantification techniques. The study's findings demonstrate that uncertainty quantification enhances transparency and provides measurable confidence in predictions, improving the explainability and clarity of black-box models. This facilitates informed decision-making and mitigates the influence of dark-patterns on user interfaces. These results highlight the importance of incorporating uncertainty quantification techniques in developing machine learning models, particularly in domains where interpretability and trustworthiness are critical.

\end{abstract}

\end{frontmatter}

\section{Introduction}

The field of NLP was revolutionized with the arrival of transformer models, a groundbreaking architecture introduced by Vaswani \textit{et al.} in their seminal work, "Attention is All You Need"~\cite{vaswani2017attention}. Prior to this, NLP relied heavily on Convolutional Neural Networks (CNN), which were useful in analyzing the spatial features of the data but lacked semantic awareness and nuances. Later, Recurrent Neural Networks (RNN) were used, which processed data sequentially and struggled but faced issues with long-range dependencies within text and stability during training due to the vanishing gradient. 

To address this, gated RNNs like Long Short-Term Memory (LSTM) were introduced, which mitigated the vanishing gradient problem but were not parallelizable and required high computational demand for training~\cite{gers2002learning}. With their unique self-attention mechanism, transformers enabled parallel processing of entire data sequences, offering a substantial leap in efficiency and effectiveness. This architecture's ability to capture complex relationships across distant parts of a text significantly enhanced performance across a myriad of NLP tasks, setting new benchmarks in machine translation, sentiment analysis, and beyond. Subsequent iterations, such as BERT~\cite{devlin2018bert} and the GPT series~\cite{gpt3_paper,Radford2019LanguageMA}, further refined and extended the transformer's capabilities, embedding it as the cornerstone of modern NLP research and applications. The transformative impact of these models is not just limited to their superior performance; they have also democratized access to high-quality NLP tools, fostering innovation and expanding the field's frontiers \cite{NLP_review_multilingual_trends_2023}.

Specifically, transformer models have been widely used for sequence classification tasks, from text sentiment analysis~\cite{xu2019bert} to DNA classification~\cite{gu2023mamba}. Despite all their advantages, transformers suffer from similar limitations to other neural network-based models, \textit{i.e.} their black-box nature that makes their understanding difficult~\cite{efficient_transformers_2022}. This aspect can be critical in tasks such as autonomous driving~\cite{Dong2021Image} or medical diagnosis~\cite{Yu2018Framing}, where there is a need to obtain a measure of certainty in the model predictions before committing to any action.

Given the black-box nature of transformer models and the critical importance of reliable predictions in high-stakes applications, integrating uncertainty quantification into these models becomes paramount~\cite{efficient_transformers_2022}. However, transformer architectures' complexity and pre-trained nature present significant challenges in modifying their internal components to accommodate uncertainty measures. As a result, focusing on the final classification head offers a practical and effective approach to introduce uncertainty quantification~\cite{bayesian_conformal_2023}. 

The interpretability and reliability of transformer-based models can be improved by integrating classification heads with predictions and measures of confidence or uncertainty. This is particularly important in applications where errors can have high costs, and understanding the model's confidence can lead to better decision-making processes~\cite{certainty_drug_discovery,lu_fair_2022}. One example of a pervasive issue requiring such knowledge is the use of dark-patterns in user interfaces. These deceptive design strategies compromise user autonomy and challenge the ethical integrity of digital services. Therefore, understanding a model's confidence can help prevent such issues and promote fair and transparent digital practices.

In this paper, to address the inherent opacity of neural networks and meet the growing demand for more transparent and trustworthy AI systems, three approaches are explored to improve the interpretability and reliability of transformer-based models for dark-pattern detection: (1) dense neural networks (DNNs), (2) Bayesian neural networks (BNNs), and (3) spectral-normalized neural Gaussian Processes (SNGPs) classification heads. We examine the performance of BNNs and SNGPs in quantifying uncertainty across various deep learning models, analyze the impact of these uncertainty quantification techniques on the performance and environmental sustainability of AI models during both training and inference phases, and explore the practical implications of uncertainty modelling through real-world examples of high and low uncertainty cases.

This research enhances the detection of dark-patterns by integrating uncertainty quantification techniques in transformer models to improve the transparency and reliability of predictions, which is crucial for applications where deceptive design practices compromise user autonomy. By providing a quantitative analysis of how these models perform, the research highlights the enhanced ability to identify dark-patterns while addressing potential impacts on environmental sustainability. The goal is to develop AI systems that are both ethically responsible and environmentally considerate, aligning with the growing demand for transparency and trust in AI applications.

\subsection{Classification Heads}

\subsubsection{Dense Neural Network layers (DNNs)}
Dense layers, also known as fully connected layers, are the most basic form of a neural network layer, where each input neuron is connected to every neuron in the next layer~\cite{KHAN2023100026}. In classification tasks, a dense layer typically serves as the final layer that maps the learned representations to the target classes. The primary advantage of dense layers lies in their simplicity and effectiveness in learning complex patterns through these direct connections. However, they do not inherently provide measures of uncertainty in their predictions, treating all inputs with equal certainty.

\subsubsection{Bayesian Neural Network layers (BNNs)}
Bayesian dense layers~\cite{jospin2022hands} extend the concept of dense layers by incorporating Bayesian inference into the network's architecture. Unlike traditional dense layers with fixed weights after training, Bayesian dense layers treat weights as distributions, This approach allows the network to simulate multiple possible models of parameters $\theta$
with an associated probability distribution $p(\theta)$, enhancing the model's ability to express and quantify uncertainty in its predictions.
This is crucial for critical applications like drug discovery and fair AI systems,  where decision-making relies heavily on the reliability of the model's output~\cite{certainty_drug_discovery,lu_fair_2022}. 

During training, BNNs utilize a prior distribution for weights instead of fixed values, reflecting initial beliefs which are updated via a likelihood function assessing the model’s fit to the data. Bayesian inference computes a posterior distribution combining these elements, often approximated through variational inference for practical implementation. After training, the BNN generates multiple predictions by randomly sampling different sets of weight values from the posterior distribution of weights. This process offers insights into into their uncertainty and impacting the predictive uncertainty. By comparing these multiple predictions, the degree of uncertainty can be assessed, where low variability among predictions indicates low uncertainty, while high variability suggests greater uncertainty. This is uncertainty quantified using multiple forward passes to generate a distribution of predictions. 

However, the major challenge with Bayesian dense layers is their computational complexity and the need for more sophisticated training techniques to manage the probabilistic nature of the weights.

\subsubsection{Spectral-normalized Neural Gaussian Process}
Spectral-normalized Neural Gaussian Process (SNGP)~\cite{liu2020simple} is a relatively recent approach that combines the ideas of spectral normalization and Gaussian Processes (GPs) with deep learning to enhance a deep classifier's capacity to measure the distance between the test example and the training data. Spectral normalization~\cite{miyato2018spectral} is a technique used to stabilize the training of neural networks by normalizing the weight matrices, ensuring that the Lipschitz constant of the function (represented by the network) is constrained. This helps maintain the model's generalization ability. On the other hand, GPs provide a principled, probabilistic approach to learning in kernel machines, offering a powerful tool for uncertainty quantification. By integrating GPs with deep learning, SNGP layers aim to preserve the deep neural network's capacity for feature extraction and representation learning while enhancing the model's ability to provide meaningful uncertainty estimates for its predictions. This makes SNGP particularly appealing for tasks requiring a careful balance between performance and interpretability, such as safety-critical applications.

\subsubsection{Motivation}
dark-patterns, defined as deceptive user interface designs that manipulate online users into unintended actions have emerged as significant threats to privacy, fairness, and transparency in digital environments. These designs exploit psychological vulnerabilities, leveraging mechanisms such as obfuscation, false urgency, and social proof to influence user decisions in favour of the companies implementing them~\cite{dark_pattern_psyco_2021}. The importance of detecting these manipulative strategies cannot be overstated, as they compromise user autonomy and undermine the ethical foundation of digital services~\cite{jarovsky_dark_2021, WALDMAN2020_dark_patter_legal}. Machine learning and deep learning algorithms have demonstrated great accuracy in identifying deceptive practices in digital ecosystems~\cite{Yada2022Dark}. This offers a solution to enhance user protection and promote transparency and fairness. However, to ensure effective detection, it is important to utilize state-of-the-art architectures and incorporate certainty in the detection process~\cite{fehr_trustworthy_2024}. Thus, combating dark-patterns is crucial, aligning with the growing demand for digital accountability and protection of consumers in complex online landscapes. This relaionship highlights the ongoing efforts and significant impact of research in tackling unethical practices in digital user interfaces.

\section{Related work}


Transformers have been widely used for text classification tasks, showcasing their versatility and efficacy across a wide array of applications. One notable application is in the domain of customer feedback analysis~\cite{nissa2023multi}, where the authors demonstrate transformers' ability to handle complex multi-label classification tasks. This research highlights how transformers, with their deep contextual understanding, can effectively categorize customer reviews into multiple relevant categories, thus providing valuable insights into customer sentiment and preferences.

The ability of transformer models to handle noisy data is crucial for maintaining their performance in real-world applications. A study titled "Transferable Post-hoc Calibration on Pretrained Transformers in Noisy Text Classification"~\cite{transformer_post_hoc} proposes post-hoc calibration techniques to fine-tune pretrained transformer models, enabling them to classify texts accurately even in the presence of noise and variability. This study demonstrates the adaptability of transformers to diverse and challenging datasets and states the importance of managing certainty in the predictions of black-box deep learning models.

Another study~\cite{ojo2023transformer} further illustrates the effectiveness of transformer-based approaches in sentiment analysis. This study explores the nuanced capabilities of transformer models in detecting sentiment, leveraging their deep learning architecture and attention mechanism to comprehend and interpret the nuances of human emotions expressed in text. The research suggests that transformer models can accurately classify texts that express a clear opinion. However, they struggle with ambiguous and ambivalent linguistic patterns. Therefore, identifying the data presented in such situations is crucial for improving the model's performance. 

Additionally, text classification can be challenging due to the lack of labelled data, which can significantly reduce model performance. Zhang \textit{et al.} \cite{TIDDI2022103627} have proposed a solution to this problem by incorporating knowledge graphs with transformational models. The research emphasises the importance of data quality. In data-scarce scenarios, it is essential to have diverse and challenging data to help the model learn and improve further. There is a need for future research in this area, particularly regarding the application of certainty to identify points where data can be strengthened.

Together, these studies paint a comprehensive picture of the state-of-the-art in the application of transformer models to text classification, highlighting their flexibility, efficiency, and effectiveness across varied contexts and challenges.

In the development of Bayesian Neural Networks (BNNs), significant strides have been made across various domains, demonstrating their versatility and effectiveness in enhancing classification tasks through uncertainty estimation. A notable advancement is presented by Bensen \textit{et al.}~\cite{bensen2023bayesian}, where a hierarchical structure within BNNs is tailored for convolutional networks. This approach capitalizes on the inherent uncertainty estimation capabilities of BNNs to improve classification outcomes, particularly in complex visual tasks. 
Similarly, Milanes \textit{et al.}~\cite{milanes2023robust} showcase the application of BNNs in the biomedical field. Here, BNNs are leveraged to classify motor imagery tasks, proving particularly adept at managing the inherent noise in electroencephalogram (EEG) data, thus underscoring the robustness of BNNs in handling data with significant variability. 
In the realm of image classification, the effectiveness of BNNs is further highlighted in~\cite{bessai2022bayesian}. 
This research emphasizes how uncertainty estimation intrinsic to BNNs can bolster prediction confidence, thereby enhancing the reliability of image classification systems. Extending this integration of Bayesian methods into convolutional neural networks (CNNs), Ferrante \textit{et al.}~\cite{ferrante2022bayesnetcnn} explore the incorporation of uncertainty directly into the network architecture. This integration not only improves the performance in image-based classification tasks but also provides a clearer understanding of the model's decision-making process, making it more interpretable.

\javi{Gaussian Processes (GPs) are a powerful, flexible and widely used Bayesian non-parametric framework for modeling and inference in a wide range of domains, from machine learning and computer science to physics, engineering, and the natural sciences \cite{rasmussen2003gaussian}. GPs are particularly well-suited for modeling complex, nonlinear, and multi-modal systems, as well as for handling small sample sizes and high-dimensional input spaces. At a high level, GPs provide a prior distribution over functions, which can be combined with data to obtain a posterior distribution over functions. This posterior distribution is also a GP, which can be used for a variety of tasks, such as function interpolation, extrapolation, optimization, and uncertainty quantification. One of the key advantages of GPs is that they provide a principled way to quantify and propagate uncertainty, which is especially important in applications where the data is noisy or the underlying system is not well understood.}
\javi{Wang \textit{et al.} \cite{wang2017educational} use GPs to focus on distinguishing educational content from non-educational materials, using word embeddings generated with Word2Vec \cite{mikolov2013efficient}. Jayashree \textit{et al.} \cite{jayashree2020evaluation} use GPs with convolutional kernels to benchmark the performance of GPs in text classification tasks for different datasets.}
Ye \textit{et al.}~\cite{Ye2023EfficientUE} use SNGPs combined with focal loss for reliable dialog response retrieval. 

Complementing the focus on certainty quantification, the principles of GreenAI underscore the need for environmentally sustainable practices in AI research and development~\cite{huertas2023comparative,schwartz2020green}. As models grow in complexity and size, substantial computational resources are required, leading to significant energy consumption and environmental impact. Aligning with the objectives of GreenAI, the ability to assess the certainty of model outputs accurately allows for more efficient allocation of computational resources. Resources can be optimized when the model exhibits high confidence, reducing unnecessary energy consumption. Conversely, in cases of high uncertainty, the model requires additional processing or data. 

The integration of certainty quantification techniques, such as BNNs or SNGPs, can enhance the interpretability and reliability of AI systems. By accurately quantifying uncertainty, these techniques can identify instances where the model is highly confident, minimizing the need for excessive computational resources and associated energy consumption. These techniques will ensure that the integration of sustainability into AI research and development aligns with pressing environmental objectives while also fostering the development of AI systems that are both reliable and energy-efficient. 

In this context, the ability to quantify uncertainty in AI predictions becomes an invaluable asset. Accurately assessing certainty can optimize computational resources, reduce energy consumption, and contribute to realizing Green AI principles, enhancing the interpretability and trustworthiness of AI systems and promoting sustainable practices in AI development.

\section{Methodology}

\javi{This section describes the components of the dark-pattern classification with uncertainty quantification task. We begin by describing the dark-patterns dataset used for this study. Then, we provide an overview of the model and head selection for the task.}

\begin{table}[t]
\centering
\renewcommand{\arraystretch}{1.5}
\begin{tabular*}{\columnwidth}{@{\extracolsep{\fill}} ll}
\hline
\textbf{Text}                                 & \textbf{Label} \\ \hline
FLASH SALE | LIMITED TIME ONLY Shop Now & Positive       \\ \hline
Write a review          & Negative       \\ \hline
Hurry! Only 2 left in stock & Positive      \\ \hline
International Shipping Policy       & Negative           \\ \hline
1142 people have added to cart recently      & Positive  \\ \hline
\end{tabular*}
\caption{Examples from the dark patterns dataset.}
\vspace{5mm}
\label{tab:text_classification_examples}
\end{table}

\subsection{Data}

\javi{The dataset used is the dark-patterns dataset developed by Yada \textit{et al.} \cite{Yada2022Dark}, containing 2356 examples scraped from different websites.}

\alvaro{It consist of a binary problem equally balanced for English language. No pre-processing is applied maintaining the case of raw text. The texts consists of examples of dark-patterns and normal patters with maximum length of 857 characters, median length 463 and min length 1 characters, with currency, Han and Hiragana symbols <1\% each.}

\subsection{Models}
The choice of models for this comparison is grounded in their innovative contributions and varied approaches to NLP and machine learning challenges:



\begin{table}[t]
\centering
\small 
\begin{tabular*}{\columnwidth}{@{\extracolsep{\fill}}ll}
\hline
\textbf{Parameter}      & \textbf{Value}                      \\ \hline
Learning Rate     & $2*10^{-5}$                           \\
Learning Rate Scheduler  & Factor of 0.1  \\
Scheduler Patience       & 1 epoch     \\
Weight Decay            & 0.01                                \\
Epochs                  & 500                                 \\
Batch Size              & 16                                  \\
Early Stopping Patience & 5                                   \\ \hline
\end{tabular*}
\caption{Training parameters for all models}
\vspace{5mm}
\label{table:training_parameters}
\end{table}

\begin{itemize}
    \item \textbf{Dolphin-Llama2-7B-AWQ}\footnote{\href{https://huggingface.co/cognitivecomputations/dolphin-llama2-7b}{https://huggingface.co/cognitivecomputations/dolphin-llama2-7b}}: An advanced model originating from the LLaMA2 architecture~\cite{touvron2023llama}, renowned for its natural language understanding and generation capabilities, especially in conversational contexts. It is enhanced by training on the Dolphin dataset\footnote{\href{https://huggingface.co/datasets/cognitivecomputations/dolphin}{https://huggingface.co/datasets/cognitivecomputations/dolphin}} to eliminate bias and alignment issues, making it particularly effective for dark-pattern detection. This model incorporates AWQ technology~\cite{lin2023awq} for 4-bit weight quantization, optimizing efficiency and speed without sacrificing accuracy, highlighting its potential for rapid and precise analysis in identifying manipulative digital interfaces. 
    \item \textbf{bert-large-uncased}: BERT~\cite{devlin2018bert} is a foundational model that significantly advanced the understanding of context in language, as the first model to successfully apply Transformers at scale. The selection of its largest variant, for our study serves a dual purpose: to benchmark the evolution of model architectures over time and to ensure a comprehensive analysis by employing the most capable version.
    \item \textbf{Mistral-7B-OpenOrca-AWQ}\footnote{\href{https://huggingface.co/TheBloke/Mistral-7B-OpenOrca-AWQ}{https://huggingface.co/TheBloke/Mistral-7B-OpenOrca-AWQ}}:  Mistral model version quantized with AWQ method trained on Q\&A OpenOrca Microsoft Dataset~\cite{OpenOrca} augmented with GPT-4 and GPT-3.5. The foundational Mistral~\cite{jiang2023mistral} model focuses on balancing computational efficiency with performance, suitable for diverse application scenarios.
    \item \textbf{mamba-370m}\footnote{\href{https://huggingface.co/state-spaces/mamba-370m-hf}{https://huggingface.co/state-spaces/mamba-370m-hf}}: Mamba~\cite{gu2023mamba} leverages new architecture based on selective state-space models (SSMs). Unlike Transformers, Mamba selectively retains or discards information based on the current token without attention, significantly reducing complexity from quadratic to linear with sequence length. This architectural innovation is especially relevant for analyzing complex webpage elements in dark-pattern detection, offering a promising approach. 
    
    \item \textbf{nomic-embed-text-v1}\footnote{\href{https://huggingface.co/nomic-ai/nomic-embed-text-v1}{https://huggingface.co/nomic-ai/nomic-embed-text-v1}}: Nomic Embed~\cite{nussbaum2024nomic} innovates in embedding techniques to provide more dynamic, context-aware representations, surpassing leading models as of February 2024.  Its top 25 ranking on the MTEB leaderboard\footnote{\href{https://huggingface.co/spaces/mteb/leaderboard}{https://huggingface.co/spaces/mteb/leaderboard}} for tasks critical to our project, like Semantic Search and Summarization, underscores its suitability for analyzing web content. With a compact size, a low memory usage footprint and advanced training methods, Nomic Embed efficiently processes up to 8192 tokens, making it ideal for identifying manipulative elements in extensive online materials. 
\end{itemize}

\alvaro{
It is important to note that all these models are open-source, contributing to the democratization of advanced artificial intelligence tools and fostering innovation across the field. 

The quantized LLM models  are quantized with Activation-aware Weight Quantization (AWQ)~\cite{lin2023awq}. AWQ focuses on low-bit weight quantization (INT3/4) recognizing that all weights are not equally important. AWQ's selective quantization preserves essential model performance while ensuring that models remain lightweight and fast, making it a key technology for the effective and efficient application of advanced AI in real-world scenarios.

Our model development strategy employs the "pre-train and fine-tune" paradigm~\cite{NIPS2015_7137debd, howard-ruder-2018-universal,kalyan2021ammus}, utilizing pre-trained models for fine-tuning. This stage is crucial for enhancing the models' performance and certainty in the downstream task of identifying dark-patterns. Through this methodology, we aim to deepen our understanding of these models' capabilities in specific real-world scenarios, focusing on the critical intersection of performance and certainty in detecting dark-patterns.
}

\begin{figure}[t]
\centering
\includegraphics[width=\linewidth]{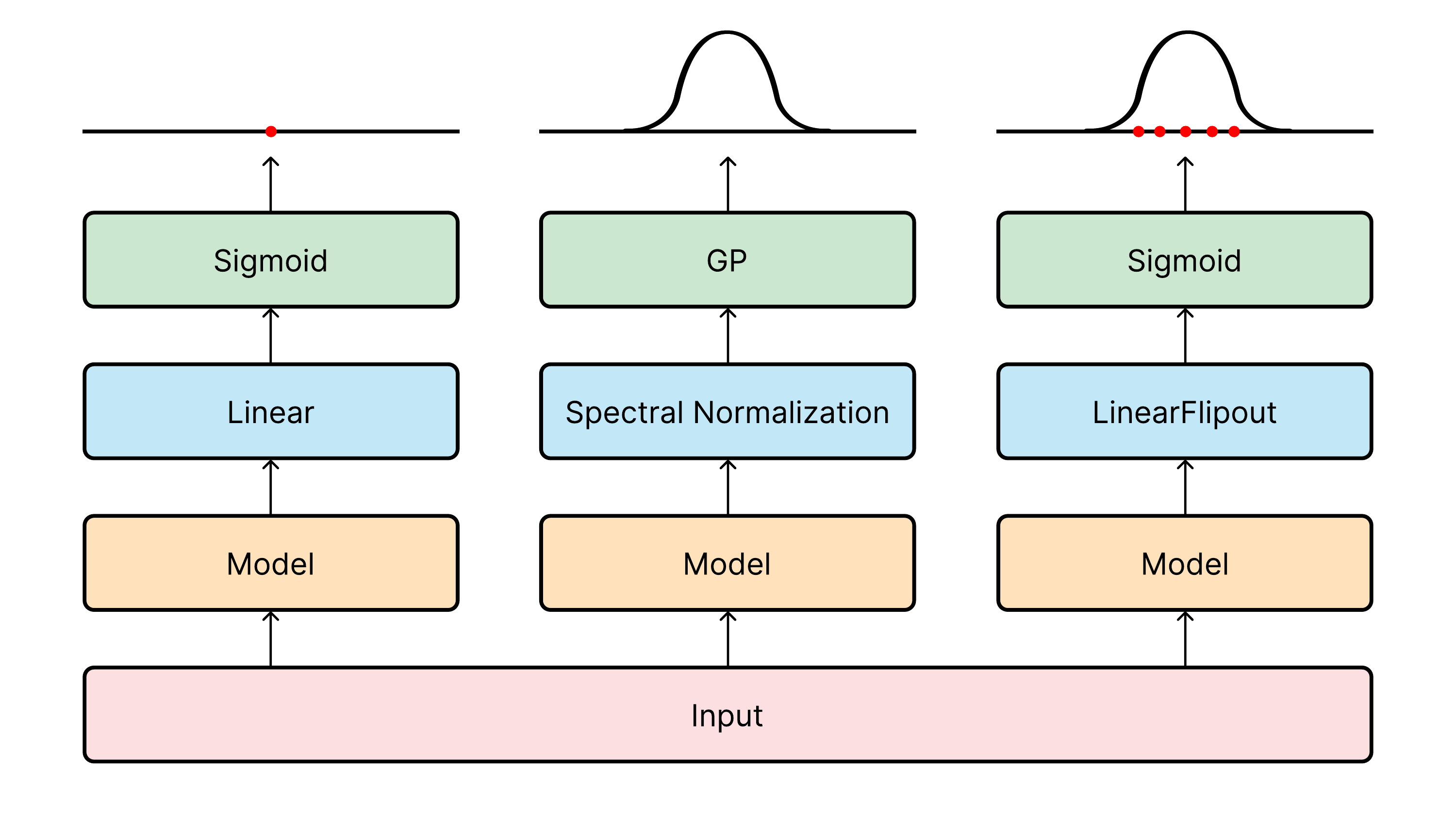}
\caption{Methodology flowchart.}
\label{fig:flow}
\vspace{18pt} 
\end{figure}

\begin{table*}[!ht]
\centering
\resizebox{\textwidth}{!}{%
\begin{tabular}{@{}ccccccccccc@{}}
\toprule
& & \textbf{Params} & \textbf{Trainable Params} & \textbf{Accuracy} & \textbf{F1} & \textbf{Inference (ms)} & \textbf{Train emissions (g)} & \textbf{Test emissions (g)} \\ \midrule
\multirow{3}{*}{BERT} & SNGP & \num{335142912} & \num{335142912} & 0.957 & 0.957 & 30.34 $\pm$ 5.73 & 1.938 & 0.031 \\
& DNN & \num{336192513} & \num{336192513} & 0.955 & 0.955 & \textbf{27.06 $\pm$ 6.41} & \textbf{1.315} & \textbf{0.03} \\
& BNN & \num{337243138} & \num{337243138} & \textbf{0.968} & \textbf{0.967} & 279.7 $\pm$ 62.5 & 1.690 & 0.312 \\ \midrule
\multirow{3}{*}{Mistral} & SNGP & \num{1196184576} & \num{1024} & 0.6165 & 0.6314 & 132.38 $\pm$ 8.74 & 46.045 & \textbf{0.278} \\
& DNN & \num{1200379905} & \num{4196353} & \textbf{0.9407} & 0.9394 & \textbf{125.42 $\pm$ 10.41} & \textbf{12.977} & 0.28 \\
& BNN & \num{1204576258} & \num{8392706} & \textbf{0.9407} & 0.\textbf{9402} & 1457.2 $\pm$ 123.1 & 14.378 & 2.836 \\ \midrule
\multirow{3}{*}{Mamba} & SNGP & \num{371517440} & \num{371517440} & 0.9004 & 0.9011 & \textbf{59.21 $\pm$ 3.08} & 16.492 & \textbf{0.062} \\
& DNN & \num{372567041} & \num{372567041} & \textbf{0.9237} & \textbf{0.9237} & 69.9 $\pm$ 9.5 & \textbf{7.058} & 0.063 \\
& BNN & \num{373617666} & \num{373617666} & 0.917 & 0.917 & 658.6 $\pm$ 94.2 & 7.304 & 0.645 \\ \midrule
\multirow{3}{*}{Nomic} & SNGP & \num{136732672} & \num{136732672} & 0.9576 & 0.9567 & 18.35 $\pm$ 2.07 & \textbf{0.514} & \textbf{0.014} \\
& DNN & \num{137520129} & \num{137520129} & 0.9555 & 0.9552 & \textbf{17.01 $\pm$ 1.56} & 1.655 & \textbf{0.014} \\
& BNN & \num{138308610} & \num{138308610} & \textbf{0.9640} & \textbf{0.9633} & 190.8 $\pm$ 30.1 & 2.135 & 0.156 \\ \midrule
\multirow{3}{*}{Llama} & SNGP & \num{1128829952} & \num{1024} & 0.887 & 0.890 & 155.49 $\pm$ 13.68& 29.842 & 0.27 \\
& DNN & \num{1133025281} & \num{4196353} & \textbf{0.921} & \textbf{0.922} & \textbf{149.69 $\pm$ 13.27} & 11.537 & \textbf{0.266} \\
& BNN & \num{1137221634} & \num{8392706} & 0.883 & 0.880 & 1711.1 $\pm$ 151.3 & \textbf{10.487} & 2.758 \\ \bottomrule
\end{tabular}
}
\caption{Combined table of CO2 emissions (g), accuracy, F1, and other metrics of all models.}
\vspace{2mm}
\label{table:results}
\end{table*}

\javi{BNNs use different methods for variance reduction, two of them being the reparametrization \cite{blundell2015weight} and flipout methods \cite{wen2018flipout}. The flipout method has emerged as a preferable variance reduction technique over the reparameterization trick. While reparameterization effectively reduces variance for models with continuous latent variables by transforming stochastic variables into deterministic functions, its application is limited to tractable distributions. Flipout, on the other hand, introduces random perturbations to gradients within mini-batches, mimicking the effects of larger batches to stabilize training without additional computational costs. This approach not only broadens its applicability, including to discrete variables and complex distributions but also reduces intra-batch interference, making it particularly suitable for the vast and varied parameter spaces of LLMs.}


\javi{In BNNs, weights are represented by probability distributions, which encode beliefs about the possible values those weights can take based on the data and prior information. To make a prediction, one must sample from these distributions, resulting in a different set of weights for each prediction. These varying sets of weights lead to a range of possible outputs for a given input, reflecting the model's uncertainty about the most appropriate weights to use. For this reason, it is necessary to compute multiple predictions for the same inputs in order to obtain a confidence interval.}

\javi{The flowchart for the methodology of this work is shown in Figure \ref{fig:flow}. The DNN, SNGP and BNN classification heads are added to each model for the dark-patterns binary classification tasks. The DNN head outputs single-point predictions while the SNGP head outputs a probability distribution and the BNN head outputs multiple predictions which form a confidence interval.}

\section{Results and Discussion}




This study extensively explores the trade-off between performance, certainty, and sustainability of Transformer models, focusing on Dense Neural Networks (DNNs) as a baseline, Bayesian Neural Networks (BNNs), and Spectral-normalized Gaussian Processes (SNGPs) for uncertainty quantification.

\javi{For the experiments, every model is fine-tuned on the dark-patterns dataset with the three different classification heads: DNN, BNN and SNGP. We compare the model size, accuracy, F1, inference time and train and test carbon emissions measured with Codecarbon~\cite{lottick2019energy}. For the model tuning, 20\% of the data is reserved as test, and the remaining 80\% is divided into 20\% validation and 80\% train. The training parameters are defined in Table \ref{table:training_parameters}.}
\javi{All experiments are conducted in a cloud server with an Nvidia RTX 5000 GPU with 16G VRAM.}

\javi{Even though the Mistral and Llama 2 models are quantized, they remain frozen during fine-tuning since the GPU does not have enough VRAM to fit all the weights. For the remaining models, all weights are fine-tuned. Regarding the hyperparameters of the classification heads, the SNGP head has 1024 inducing points, based on the default value on the original paper \cite{liu2020simple}. For fairness of results, the DNN and BNN classification heads have a hidden layer of 1024 units, with the number of output neurons of the base model as inputs and one output neuron for the binary classification logits. For the BNN, the weight initialization parameters are $\mu = 0$ and $\sigma = 1$ and the number of predictions used for obtaining the confidence interval is 10.}

\javi{It is important to note that the main objective of this paper is not to improve the baselines on dark-pattern detection, but to use the dark-pattern classification task as a way to compare different classification heads for uncertainty quantification in transformer models.}

\begin{figure}[t]
    \centering
    \begin{minipage}{0.48\textwidth}
        \centering
        \includegraphics[width=\linewidth]{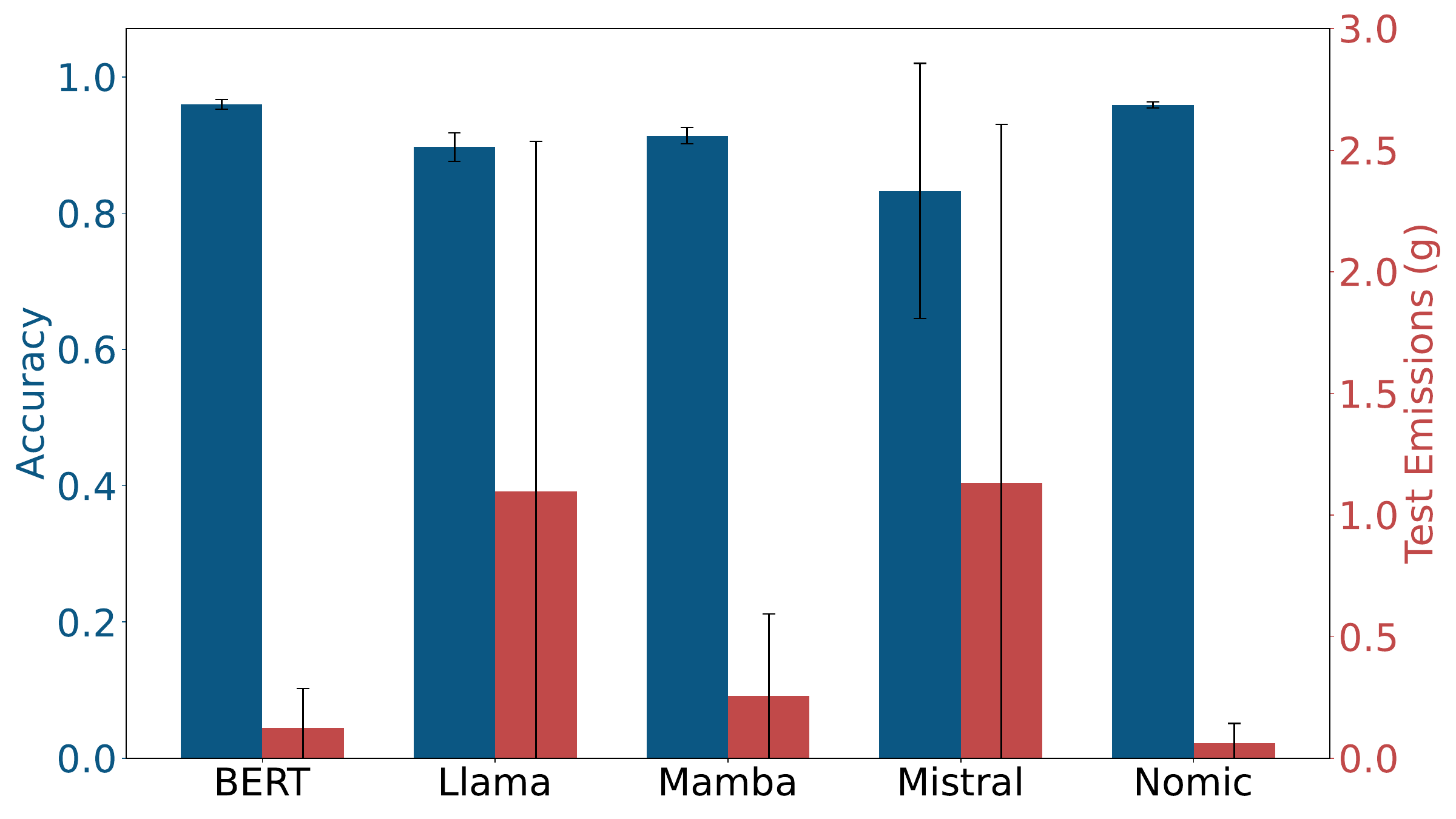}
        \caption{Accuracy vs test emissions for each model averaging the three classification heads.}
        \label{fig:model}
    \end{minipage}\hfill 
    \vspace{5mm}
    \begin{minipage}{0.48\textwidth}
        \centering
        \includegraphics[width=\linewidth]{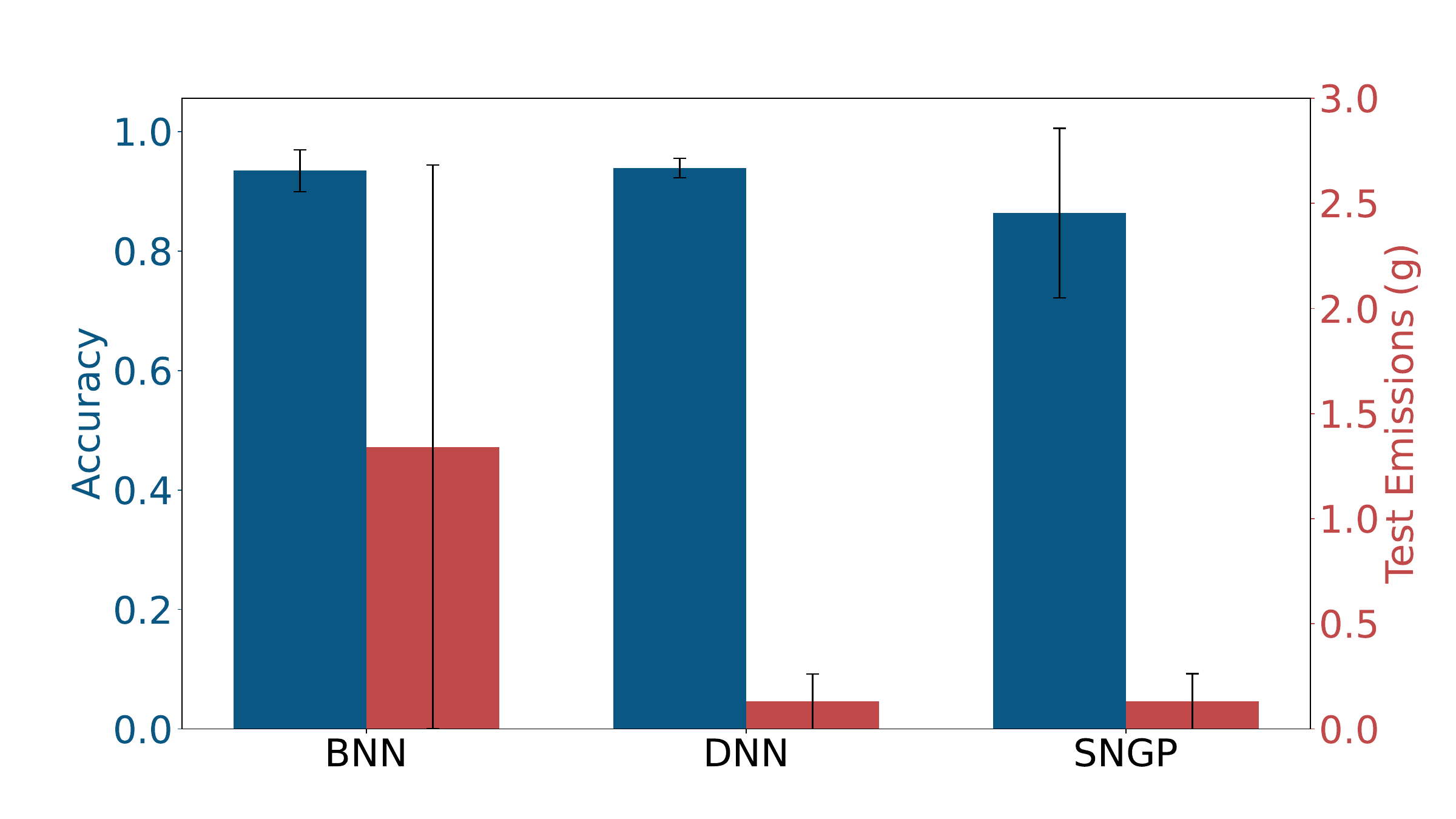}
        \caption{Accuracy vs test emissions for each classification head averaging all models.}
        \vspace{10mm}
        \label{fig:head}
    \end{minipage}
\end{figure}
 
\subsection{Trade-off Performance Analysis}
After evaluating the performance of different types of models used in this study, we discovered that each model has its unique advantages and limitations based on specific accuracy and inference time metrics (see Table~\ref{table:results}). These performance metrics significantly impact determining the suitability of each model type for various practical applications.

DNNs displayed consistent accuracy across all tests, making them a reliable choice for applications where stable and predictable performance is needed. Due to their simpler architecture, DNNs also had the fastest inference times among the models tested. This makes them particularly valuable in scenarios where rapid decision-making is critical, such as in real-time systems where delay can result in inefficiencies or safety concerns.

BNNs, while offering the advantage of quantifying uncertainty, showed variability in performance, particularly in accuracy. This variability originates from their probabilistic nature, which, while providing a deeper insight into the model's confidence levels, can lead to less predictability in outcomes. The inference times for BNNs were significantly longer than those for DNNs due to the computational overhead of managing probabilistic weights and multiple sampling to estimate uncertainty. This model type is best suited for applications where prediction confidence is more critical than prediction speed, such as in strategic decision-making environments where incorrect decisions could have high consequences.

In the case of DNNs vs BNNs, DNNs have half the number of parameters of BNNs, as BNNs require a mean and standard deviation for each weight, and can be trained easily with a cross entropy loss, while probabilistic models usually use Kullback-Leibler divergence, which is not as straigthforward. In the case of SNGPs, they use a simpler Laplace approximation to the original Gaussian process, as exact and even approximate GPs suffer from a high computational cost as they need to invert the covariance matrix to obtain the predictive Gaussian distribution. DNNs are also a good choice when the inference time requirements don't allow for the multiple predictions of BNNs or the slightly slower predictions of SNGPs.

SNGPs were designed to balance accuracy and the quantification of uncertainty. However, their integration with larger models like Mistral resulted in reduced performance and higher variability in accuracy, as indicated in Figure \ref{fig:head}. Despite this, SNGPs maintained moderate inference times and offered valuable insights into the uncertainty of predictions, making them suitable for use in environments where reliability and detailed probabilistic understanding are necessary but where the extreme computational demands of BNNs are prohibitive.

One of the most significant findings from the performance analysis was the impact of model size on accuracy and inference times. Larger models such as Llama and Mistral had slower inference times and also showed decreased accuracy in the case of Mistral using SNGP.  This result indicates the greater challenge of modelling certainty in larger models. This aspect is crucial when selecting a model for practical applications, as the benefits of larger, more complex models must be weighed against the increased resource demands and potential decrease in performance.


\begin{table}[t]
\centering
\small 
\begin{tabular*}{\columnwidth}{@{\extracolsep{\fill}}ccccc}
\toprule
                         &      & Top 10\% & Bot 10\% & Mean variance             \\ \midrule
\multirow{2}{*}{BERT}    & SNGP & 95.745   & 95.745   & 0.005                     \\
                         & BNN  & 74.468   & 100.000  & 0.052                     \\ \midrule
\multirow{2}{*}{Mistral} & SNGP & 68.085   & 57.447   & 0.931                     \\
                         & BNN  & 68.085   & 100.000  & 0.029                     \\ \midrule
\multirow{2}{*}{Mamba}   & SNGP & 74.468   & 89.362   & 0.005                     \\
                         & BNN  & 63.830   & 97.872   & 0.027                     \\ \midrule
\multirow{2}{*}{Nomic}   & SNGP & 97.872   & 97.872   & 0.005                     \\
                         & BNN  & 74.468   & 100.000  & 0.006                     \\ \midrule
\multirow{2}{*}{Llama}   & SNGP & 95.745   & 95.745   & 0.005                     \\
                         & BNN  & 63.830   & 100.000  & 0.036                     \\ \bottomrule
\end{tabular*}
\caption{Accuracy of all models on top 10\% (more uncertainty) and bottom 10\% (less uncertainty) variances.}
\vspace{5mm}
\label{table:acc_var_easy_hard_samples}
\end{table}

\subsection{Sustainability and Environmental Impact}
The study demonstrates a direct correlation between the size of the Transformer models and their carbon emissions, even when LLMs have frozen layers except for the head, such as Mistral and Llama. These larger models require significantly more computational power, translating to higher energy consumption and increased carbon emissions, as depicted in Figure~\ref{fig:model}. This relationship is crucial for organizations that balance performance with sustainability, as opting for smaller, more efficient models like Nomic could significantly reduce their environmental impact.

The environmental impact varies significantly across different model types during the training and inference phases. Traditional DNNs, while less computationally intensive than BNNs or SNGPs, do not offer uncertainty quantification, which might necessitate retraining or additional computational overhead in uncertain scenarios. Although BNNs provide valuable insights into model certainty, they also have a much higher energy cost due to the need to process multiple samples to estimate uncertainty. This is evident from the study's findings, where BNNs consumed up to ten times more energy than DNNs under similar conditions.

\subsection{Comparing Uncertainty Quantification Approaches}


\begin{table}[!t]
\centering
\small 
\begin{tabular*}{\columnwidth}{@{\extracolsep{\fill}}lc@{}}
\toprule
\multicolumn{1}{c}{Text} & Uncertainty \\ \midrule
Help us \#savethefishies  & Low \\
6 customers have this in their basket  & Low \\
Only a few more left!  & Low \\
Hurry! Limited Quantity Available.  & Low \\ \midrule
Arthritis Aids  & High \\
The presence of flowers is enough to ...  & High \\
Irwin  & High \\
tedpullin  & High \\ \bottomrule
\end{tabular*}
\caption{Predictions from the Nomic SNGP model with high and low uncertainty.}
\vspace{15pt}
\label{table: uncertainty}
\end{table}

Through the detailed review of model performance, stability, and energy consumption, we can better understand the strengths and limitations of each BNN and SNGP approach.

As shown in Table~\ref{table:acc_var_easy_hard_samples}, the analysis reveals significant differences in certainty and accuracy between SNGP and BNN models. For instance, BERT and Llama models equipped with SNGP heads showed exceptional stability with identical scores of 95.745 for the top and bottom 10\% of predictions, accompanied by a near-zero mean variance (0.005). This indicates highly stable predictions across the dataset, suggesting that SNGP models maintain consistent performance even in varying data conditions. In contrast, the Mistral model with an SNGP head displayed less stability, with a notable discrepancy between the highest and lowest 10\% of predictions (68.085 vs 57.447) and a significant mean variance (0.931).

When coupled with an SNGP, Nomic showcased high certainty, as both the top and bottom 10\% of predictions scored very high (97.872), coupled with low variance (0.005). This reflects a robust model architecture or particularly effective training data, emphasizing the potential of SNGP to provide reliable and consistent outputs.

In general, SNGP models tend to show lower mean variances than their BNN counterparts for the same set of models, indicating more stable predictions. While BNNs may occasionally reach higher peaks of certainty or confidence in certain predictions, their performance across the dataset is comparatively less stable, marked by higher mean variances. This variability could influence the choice between SNGP and BNN depending on the need for stability in application outputs.

From an energy consumption perspective, the SNGP models generally incur similar costs to traditional models without uncertainty quantification heads, whereas BNNs can be significantly more energy-intensive. For example, the BNN classification head's test emissions were ten times higher than those of DNNs. This higher energy demand is primarily because BNNs require multiple samples to measure certainty effectively. Note that BNN models could require even more computational resources if more points are needed to enhance certainty and reliability, potentially affecting their scalability and practical application in energy-sensitive environments.

The use of an SNGP head implies an additional energy cost of approximately 1.2\% on average, which is minimal compared to the substantial ten times increase associated with BNN heads. This analysis suggests that while SNGP provides a balance between performance uncertainty stability and energy efficiency, BNNs pose challenges regarding higher operational costs and environmental impact.

\subsection{Practical Implications}

One of the study's objectives is the model's ability to quantify uncertainty in its predictions. Table
\ref{table: uncertainty} presents high and low uncertainty instances from the Nomic SNGP model. The semantic clarity of the text examples is evident in the low uncertainty cases, where the model's predictions are made with confidence. These examples are straightforward, containing complete phrases that convey a clear message, related to consumer behaviour or stock levels, such as "\textit{Only a few more left!}" or "\textit{Hurry! Limited Quantity Available.}". 

On the other hand, the high uncertainty cases consist of single words like "\textit{Arthritis Aids}" or "\textit{Irwin}", which are semantically ambiguous without further context. This ambiguity translates into a higher variance in the model's confidence. The presence of these high-variance cases in the table serves a critical function; it underscores the capability of the Nomic SNGP model to introspect and evaluate its certainty. 

This model's "self-awareness" has practical implications that add a layer of interpretability to the AI's decision-making process compared to the traditional dense head layer model counterparts. This interpretability is crucial for end-users, enabling them to discern when a model's output is reliable and when it should be treated with scepticism. This discernment is critical when considering the identification of dark-patterns. The certainty measure allows the model to signal which cases are clear-cut and ambiguous, thus avoiding the pitfall of overgeneralizing or making unwarranted assumptions based on uncertain predictions, a real-world relevance that should resonate with data scientists and AI developers.

It is important to note that the ability to measure certainty is a tool that end-users could use, and it also plays a crucial role in guiding the development of the model itself. As discussed in a recent study by Schmarje \textit{et al.}~\cite{kothe_label_2024}, having high-quality data and addressing label ambiguity is crucial for data scientists to identify gaps in the training data or areas where the model require further improvement. Knowing the level of certainty in the model's predictions can significantly improve the learning process by analyzing instances of high uncertainty that indicate the model's difficulties and which predictions need reevaluation or additional context.

\section{Conclusions and Future Work}

This research paper focuses on enhancing the interpretability of transformer models by integrating uncertainty quantification, aimed specifically at detecting dark-patterns in user interfaces as an example of risky situations where certainty is valuable. We demonstrate that this approach can make AI systems more trustworthy without significantly compromising performance. Our study uses dense layers, Bayesian dense layers, and spectral-normalized neural Gaussian processes to achieve this goal. The evaluations across various metrics—model size, accuracy, inference time, and environmental impact—indicate that while there are trade-offs, particularly regarding computational demand and carbon footprint, the benefits of increased reliability and accountability in model predictions are profound. 
While we have made progress, there are still areas to explore, particularly in detecting and mitigating bias in text-based AI applications, where dark-patterns can skew outcomes unfavourably. We suggest that uncertainty quantification methods can be adapted to identify and correct biases in training data or model predictions. Additionally, we propose using conformal prediction and distance awareness to establish confidence intervals around predictions, providing a clear statistical guarantee about their accuracy. Applying these methods to transformer models can further enhance their usability in risk-sensitive environments.



\begin{ack}
\javi{The funding for this work was provided by Funditec.}
\end{ack}



\bibliography{mybibfile}

\end{document}